\documentclass[conference]{IEEEtran}
\IEEEoverridecommandlockouts
\usepackage{tikz}
\usepackage{textcomp}
\usepackage{hyperref}
\usepackage{lipsum}

\newcommand\copyrighttext{%
  \footnotesize \textcopyright 2022 IEEE. Personal use of this material is permitted.
  Permission from IEEE must be obtained for all other uses, in any current or future
  media, including reprinting/republishing this material for advertising or promotional
  purposes, creating new collective works, for resale or redistribution to servers or
  lists, or reuse of any copyrighted component of this work in other works.
  DOI: \href{https://ieeexplore.ieee.org/document/9751184}{1109/CISS53076.2022.9751184}}
\newcommand\copyrightnotice{%
\begin{tikzpicture}[remember picture,overlay]
\node[anchor=south,yshift=10pt] at (current page.south) {\fbox{\parbox{\dimexpr\textwidth-\fboxsep-\fboxrule\relax}{\copyrighttext}}};
\end{tikzpicture}%
}

\usepackage{cite}
\usepackage{amsmath,amsfonts,bm}
\usepackage{amsthm}
\usepackage{amssymb}
\usepackage{algorithm}
\usepackage{graphicx}
\usepackage{subfigure}

\newtheorem{thm}{Theorem}
\newtheorem{definition}{Definition}
\newtheorem{corollary}{Corollary}

\def\bfe{{\boldsymbol e}}

\def\bfv{{\boldsymbol v}}
\def\bfw{{\boldsymbol w}}
\def\bfx{{\boldsymbol x}}

\def\bfz{{\boldsymbol z}}
\def\bfmu{{\boldsymbol \mu}}

\def\bfSg{{\boldsymbol\Sigma}}

\def\bfA{{\boldsymbol A}}
\def\bfB{{\boldsymbol B}}
\def\bfC{{\boldsymbol C}}

\def\bfI{{\boldsymbol I}}

\def\bfM{{\boldsymbol M}}

\def\bfP{{\boldsymbol P}}
\def\bfQ{{\boldsymbol Q}}
\def\bfR{{\boldsymbol R}}

\def\bfT{{\boldsymbol T}}
\def\bfU{{\boldsymbol U}}

\def\bfW{{\boldsymbol W}}

\def\bfZ{{\boldsymbol Z}}

\newcommand{\R}{\mathbb{R}}

\newcommand{\be}{\begin{equation}}
\newcommand{\ee}{\end{equation}}

\def\eqref#1{(\ref{#1})}
\def\1{\bm{1}}

\DeclareMathAlphabet{\mathsfit}{\encodingdefault}{\sfdefault}{m}{sl}
\SetMathAlphabet{\mathsfit}{bold}{\encodingdefault}{\sfdefault}{bx}{n}

\usepackage{amsmath,amssymb,amsfonts}
\usepackage{algorithmic}
\usepackage{graphicx}
\usepackage{textcomp}
\usepackage{xcolor}
\def\BibTeX{{\rm B\kern-.05em{\sc i\kern-.025em b}\kern-.08em
    T\kern-.1667em\lower.7ex\hbox{E}\kern-.125emX}}
\begin{document}

\title{Learning and generalization of one-hidden-layer neural networks, going beyond standard Gaussian data\\
\thanks{This work was supported in part by AFOSR FA9550-20-1-0122,  ARO W911NF-21-1-0255, NSF 1932196, and the Rensselaer-IBM AI Research Collaboration (http://airc.rpi.edu), part of the IBM AI Horizons Network (http://ibm.biz/AIHorizons).}
}

\author{\IEEEauthorblockN{ Hongkang Li}
\IEEEauthorblockA{
% \textit{Dept. of Electrical, Computer, and Systems Engineering} \\
\textit{Rensselaer Polytechnic Institute}\\
Troy, NY, USA\\
lih35@rpi.edu}
\and
\IEEEauthorblockN{ Shuai Zhang}
\IEEEauthorblockA{
% \textit{Dept. of Electrical, Computer, and Systems Engineering} \\
\textit{Rensselaer Polytechnic Institute}\\
Troy, NY, USA\\
zhangs21@rpi.edu}
\and
\IEEEauthorblockN{ Meng Wang}
\IEEEauthorblockA{
% \textit{Dept. of Electrical, Computer, and Systems Engineering} \\
\textit{Rensselaer Polytechnic Institute}\\
Troy, NY,  USA\\
wangm7@rpi.edu}
}

\maketitle
\copyrightnotice
\begin{abstract}
This paper analyzes the convergence and generalization of training a one-hidden-layer neural network when the input features follow the Gaussian mixture model consisting of a finite number of Gaussian distributions. Assuming the labels are generated from a teacher model with an unknown ground truth weight, the learning problem is to estimate the underlying teacher model by minimizing a non-convex risk function over a student neural network. With a finite number of training samples, referred to the sample complexity,  the iterations are proved to converge linearly to a critical point with guaranteed generalization error. In addition, for the first time, this paper characterizes the impact of the input distributions on the sample complexity and the learning rate.
\end{abstract}

\begin{IEEEkeywords}
	Gaussian mixture model, convergence, generalization, sample complexity, neural networks
\end{IEEEkeywords}

\section{Introduction}
The recent success of machine learning mainly benefits from the developments of neural networks. It is surprising to witness the superior empirical performance in various applications ranging from imaging processing \cite{KSH12,HZRS16, KCKJ22} to natural language processing \cite{GMH13}. However, the lack of theoretical generalization guarantee has become a rising concern with the widespread utilization of neural networks. That is,  the learned model from a finite number of training samples has guaranteed test error on the unseen data. From the perspective of optimization, achieving guaranteed generalization is to obtain both a small training error and a bounded generalization gap (difference between the training and test errors), which are referred to convergence and generalization analysis. Despite the high test accuracy in numerical experiments, the convergence and generalization analysis are limited, and significant breakthroughs focus on shallow neural networks \cite{MMN18,JGH18,ZSJB17}. Nevertheless, it is well-known that training a one-hidden-layer neural network with only three nodes is NP-hard \cite{BR92}, and various assumptions are imposed to ensure feasible analysis.

One representative line of works studies the overparameterized neural networks, where the number of trainable parameters is far larger than the number of training samples. In particular, the optimization problem of training on overparameterized networks has no spurious local minima \cite{LSS14,LY17,ZBHRV16,SJL18}, and thus the random initialization can achieve bounded training errors. However, such learned models may suffer from over-fitting and cannot provide a generalization guarantee without additional assumptions. With the assumptions of infinite width of neurons, training a neural network via gradient descent can be approximated by a differential equation from the Mean Field (MF) view \cite{CB18b,MMN18,FGZZ19,Nguyen2019} or a Gaussian kernel processing from the Neural Tangent Kernel (NTK) view \cite{JGH18,ALS19,DZPS19,ZCZG20,ZG19, LWLC22}. In particular,  for one-hidden-layer neural networks,  \cite{MMM19,WLLM18,ALL19,ADHL19}
provide bounded generalization errors for some target functions with nice properties, i.e., even or infinitely smooth functions. Nevertheless, both MF and NTK frameworks require a significantly larger number of neurons  than that in practical cases.

For training neural networks with a fixed number of neurons, the landscapes of the objective functions are highly non-convex. They can contain intractably  many spurious local minima \cite{SS17} with multiple neurons.
This line of work follows the ``teacher-student'' setup, where the training data is generated by a teacher neural network. The learning is performed on a student network by minimizing the empirical risk functions of the training data.   
To achieve the optimal with guaranteed generalization, most of the theoretical works  with a fixed number of neurons assume the input belongs to standard Gaussian distributions (zero mean and unit variance) for feasible analysis \cite{ZSJB17,ZWXL20,ZWLC20_1,ZWLC20_2,ZWLC21,FCL20} with a few exceptions \cite{DLT18,MBM16}.
Nevertheless, the neural network considered in \cite{DLT18,MBM16} only contains a single neuron in the hidden layer, and the objective function has a large benign landscape near the desired point such that random initialization succeeds with constant probability regardless of the non-convexity. 
Other works considering non-Gaussian input features either require an infinitely large number of neurons \cite{MMN18,LL18} or cannot provide bounded generalization analysis \cite{YO19,GMKZ19,GRMK20,MKUZ20,GMMM20}. 

Overall, the generalization analysis of neural networks beyond the standard Gaussian input is  less investigated and has not received enough attention. On the one hand, Gaussian mixture models are widely employed in plenty of applications, including data clustering \cite{D99, FJ02, J10}, image segmentation \cite{PFJ06}, and few-shot learning \cite{YLX21}. 
On the other hand, 
the study of Gaussian mixture models casts an insight into understanding how the mean and variance of the input features affect the performance. In practice, the input distribution will affect the learning performance, and several data pre-processing techniques such as data whitened \cite{LBOM98} and batch normalization \cite{IS15} have been applied to accelerate the convergence rate and improve the generalization error.
Although various works \cite{BGSW18,CPM20,STIM18} have studied the enormous success of batch normalization from different scopes and provided different explanations,  none of them have studied from the perspective of generalization analysis. 
% the learning performance clearly depends on the input data distribution. \cite{LBOM98} states that the learning method converges faster if the inputs are whitened to be the standard Gaussian. Batch normalization \cite{IS15} modifies the mean and variance in each layer   and is a popular practical method to achieve fast and stable convergence. 

% The parameters of the mixture model can be estimated from data  by   the EM algorithm \cite{RW84} or the moment-based method \cite{HK13}, with theoretical performance guarantees, see, e.g., \cite{HN16,HDKW20,DHKJ20,DHKw20}.
\textbf{Contributions:} This paper provides a  theoretical analysis of learning one-hidden-layer neural networks when the input distribution   follows a Gaussian mixture model containing an arbitrary number of Gaussian distributions with arbitrary mean. Specific contributions include: 

1. \textbf{Proposition of a gradient descent algorithm and tensor initialization with Gaussian mixture inputs:} This is the first paper to propose a gradient descent with tensor initialization method for Gaussian mixture inputs.

2. \textbf{Guaranteed generalization error with Gaussian mixture inputs for binary classification problems:} For the first time, we give a guaranteed generalization error for binary classification problems when the inputs follow Gaussian mixture model.

3. \textbf{First sample complexity analysis depending on the parameters of the input distribution:} Our theorem provides the first explicit characterization of the required number of samples to learn the neural network. 
% The proposed algorithm enjoys a linear convergence rate, and the returned critical point are proved to approach the teacher model  in the order of $\sqrt{d\log{n}/n}$, where $d$ is the dimension of the feature, and $n$ is the number of samples. 

% We also characterize  the required number of samples for accurate estimation, referred to as the sample complexity, 
% as a function of  $d$, the number of neurons $K$, and the input distribution.  Our explicit bounds imply (1)  when the absolute value of any mean   in the Gaussian mixture model increases from zero, the sample complexity increases, and the algorithm converges slower, indicating that it will be more challenging to learn a model with a small test error; (2) The same phenomenon happens when any variance in the mixture model increases to infinity from a certain positive value, or if all the variances in the mixture model   approach  zero. Our results indicate that the training converges faster and requires fewer samples if the input data are zero mean with a certain non-zero variance. This can be viewed as one theoretical explanation in one-hidden-layer for the success of Batch normalization.  Moreover, to the best of our knowledge, \textit{this paper provides the first theoretical and explicit characterization about how the mean and variance of the  input distribution affect  the sample complexity and learning rate}. 

\textbf{Notations}: Vectors are in bold lowercase, and matrices and tensors
are in bold uppercase. Scalars are in normal fonts. Sets are in calligraphic fonts. For example, $\bfZ$ is a matrix, and $\bfz$ is a vector. $\mathcal{Z}$ is a set. $z_i$ denotes the $i$-th entry of $\bfz$, and $Z_{i,j}$ denotes the $(i,j)$-th entry of $\bfZ$. 
$[K]$ ($K>0$) denotes the set including integers from $1$ to $K$. $\bfI_d\in\mathbb{R}^{d\times d}$ and $\bfe_i$ represent the identity matrix in $\mathbb{R}^{d \times d}$ and the $i$-th standard basis vector, respectively. 
We use $\delta_i(\bfZ)$ to denote the $i$-th largest singular value of $\bfZ$. The matrix norm is defined as $\|\bfZ\|=\delta_1(\bfZ)$. The gradient and the Hessian of a function $f(\bfW)$ are  denoted by $\nabla f(\bfW)$ and $\nabla^2 f(\bfW)$, respectively.\\ %The outer product of  vectors $\bfz_i\in\mathbb{R}^{n_i}$, $i\in[l]$, is defined as $\bfT=\bfz_1\otimes \cdots\times \bfz_l\in\mathbb{R}^{n_1\times\cdots\times n_l}$ with $\bfT_{j_1\cdots j_l}=(\bfz_1)_{j_1}\cdots(\bfz_l)_{j_l}$.\\
Given a tensor $\bfT\in\mathbb{R}^{n_1\times n_2\times n_3}$ and   matrices $\bfA\in\mathbb{R}^{n_1\times d_1}$, $\bfB\in\mathbb{R}^{n_2\times d_2}$, $\bfC\in\mathbb{R}^{n_3\times d_3}$, the $(i_1,i_2, i_3)$-th entry of the tensor $\bfT(\bfA,\bfB, \bfC)$ is given by
\begin{equation}
    \sum_{i_1'}^{n_1}\sum_{i_2'}^{n_2}\sum_{i_3'}^{n_3}\bfT_{i_1',i_2', i_3'}{\bfA}_{i_1',i_1}{\bfB}_{i_2',i_2}{\bfC}_{i_3',i_3}.\label{T(A,B,C)}
\end{equation}
We follow the convention that $f(x)=O(g(x))$ (or $\Omega(g(x))$, $\Theta(g(x)))$ means that $f(x)$ is at most, at least, or in the order of $g(x)$, respectively. 
\section{Problem Formulation}
In this paper, we consider the distribution $(\mathcal{X}, \mathcal{Y})$ over $(\bfx, y)\in \mathbb{R}^{d}\times \{+1, -1\}$, where $\mathcal{X}$ is the Gaussian mixture distribution \cite{P94, TSM85, HK13} such that 
\begin{equation}
\bfx \sim \sum_{l=1}^L\lambda_l\mathcal{N}(\bfmu_l,\bfI_d),\label{notation_GMM}
\end{equation}
where $\lambda_l\in[0,1]$ with $\sum_l\lambda_l=1$, and $\mathcal{N}$ denotes the multi-variate Gaussian distribution with mean  $\bfmu_l\in \R^{d}$, and covariance $\bfI_d$ for all $l\in [L]$. Let $\bfM=(\bfmu_1,\bfmu_2,\cdots,\bfmu_L)$ and $\boldsymbol{\lambda}=(\lambda_1,\lambda_2,\cdots,\lambda_L)$. 
The teacher model in this paper is a fully connected neural network consisting of an input layer, a hidden layer, followed by a pooling layer. The number of neurons in the hidden layer is denoted as $K$. All neurons are equipped with Sigmoid activation functions\footnote{The results can be generalized to any even  activation  function $\phi$  with bounded  $\phi$, $\phi'$ and $\phi''$. Examples include $\tanh$ and $\text{erf}$.} $\phi(x)=\frac{1}{1+\exp(-x)}$. For any input $\bfx \in \mathcal{X}$, the output of the teacher model is denoted as 
\begin{equation} \label{cla_model}
H(\bfW^*;\bfx):=\frac{1}{K}\sum_{j=1}^K\phi({\bfw^*_j}^\top\bfx),
\end{equation}
where $\bfW^*=[\bfw^*_1,...,\bfw^*_K]\in \R^{d\times K}$ is an unknown fixed weight matrix, and $\bfw^*_j \in \R^{d}$  is the weight of the $j$-th neuron in the hidden layer. The corresponding output label $y$  satisfies
\begin{equation}
    \mathbb{P}(y=1|\bfx)=H(\bfW^*;\bfx).
\end{equation}
The training process is over a student neural network with the same architecture as the teacher model, and the trainable parameters are denoted as $\bfW\in \R^{d\times K}$. 
Given $n$ pairs of training samples $\{\bfx_i,y_i\}_{i=1}^n$ from $(\mathcal{X},\mathcal{Y})$, the empirical risk function is 
\begin{equation}\label{empirical}
f_n(\bfW)=\frac{1}{n}\sum_{i=1}^n \ell(\bfW;\bfx_i,y_i),
\end{equation}
and $\ell(\bfW;\cdot,\cdot)$ is the cross-entropy loss function as
\begin{equation}\label{cross-entropy}
\begin{aligned}
\ell(\bfW;\bfx,y)
=&-y\cdot\log(H(\bfW;\bfx))\\
&-(1-y)\cdot\log(1-H(\bfW;\bfx)).
\end{aligned}
\end{equation}
The neural network training problem is to minimize the following non-convex objective function:
\begin{equation}\label{eqn:problem} 
\min_{\bfW \in \R^{d\times K}}: \quad  f_n(\bfW).    
\end{equation}
$\bfW^*$ may not be a global optimal solution to \eqref{eqn:problem} because $y$ is the quantization of $H(\bfW^*;\bfx)$. However, $\bfW^*$ minimizes the expectation of \eqref{eqn:problem} over $(\mathcal{X},\mathcal{Y})$.
Note that for any permutation matrix $\bfP\in\mathbb{R}^{K\times K}$, we have  $H(\bfW;\bfx)=H(\bfW \bfP;\bfx)$, and $f_n(\bfW \bfP)=f_n(\bfW)$. Hence, the estimation is considered successful if one finds weights $\bfW$ close to any column permutation of $\bfW^*$. 
\section{Algorithm}
We propose a gradient descent algorithm with tensor initialization method to solve \eqref{eqn:problem}. The method starts from an initialization $\bfW_0 \in \R^{d\times K}$ computed from the tensor initialization method (Subroutine \ref{TensorInitialization}) and then updates the iteration $\bfW_t$ using gradient descent   with the step size $\eta_0$. The pseudo-code is summarized in Algorithm \ref{gd}.

The tensor initialization algorithm is summarized in Subroutine \ref{TensorInitialization}. While existing tensor decomposition methods in \cite{JSA14} and \cite{ZSJB17} only apply for standard Gaussian distribution, we extend the methods to Gaussian mixture models. The intuition is similar to that in \cite{ZSJB17}, where the directions of $\{\bfw_j^*\}_{j\in[K]}$ are first estimated through decomposing the high-order tensor, and then the corresponding magnitudes are estimated through solving a linear regression problem. However, the high-order tensors are defined in a fairly different way because of the difference between input distributions. Formally, the high order tensor $\bfQ_j$'s are defined in Definition \ref{def: M}.
\begin{definition}\label{def: M}
 For $j=1,2,3$, we define 
\begin{equation}\label{eqn:mj}
    \bfQ_j:=\mathbb{E}_{\bfx\sim\sum_{l=1}^L\lambda_l\mathcal{N}(\bfmu_l,\bfSg_l)}[y\cdot (-1)^j p^{-1}(\bfx)\nabla^{(j)}p(\bfx)], 
 %   \bfM_j:=\mathbb{E}_{\bfx\sim\sum_{l=1}^L\lambda_l\mathcal{N}(\bfmu_l,\sigma_l^2\bfI)}[y\cdot (-1)^j\frac{\nabla^{(m)}p(\bfx)}{p(\bfx)}],\ j=1,2,3  
\end{equation}
where $p(\bfx)$ is
\begin{equation}\label{eqn:p}
 p(\boldsymbol{x})=\sum_{l=1}^L\lambda_l(2\pi)^{-\frac{d}{2}}\exp\big(-\frac{1}{2}(\boldsymbol{x}-\boldsymbol{\mu}_l)^\top(\boldsymbol{x}-\boldsymbol{\mu}_l)\big).
 \end{equation}
\end{definition}

\begin{algorithm}
\caption{Gradient Descent with Tensor Initialization}\label{gd}
\begin{algorithmic}[1]
\STATE{\textbf{Input: }} 
 Training data $\{(\bfx_i,y_i)\}_{i=1}^n$, the step size $\eta_0= \big( \sum_{l=1}^L\lambda_l (\|{\bfmu}_l\|_\infty+1)^2 \big)^{-1}$;
\STATE{\textbf{Initialization: }}$\bfW_{0}\leftarrow$ Tensor initialization method via Subroutine \ref{TensorInitialization};
\FOR{$t=0,1,\cdots,T-1$}
\item $\bfW_{t+1}= \bfW_t-\eta_0\nabla f_n(\bfW)$
\ENDFOR
\STATE{\textbf{Output: }} $\bfW_T$
\end{algorithmic}
\end{algorithm}

\floatname{algorithm}{Subroutine}
\setcounter{algorithm}{0}
\begin{algorithm}
\caption{Tensor Initialization Method}\label{TensorInitialization}
\begin{algorithmic}[1]
\STATE{\textbf{Input: }} Partition $\{(\bfx_i,y_i)\}_{i=1}^n$ into three disjoint subsets $\mathcal{D}_1$, $\mathcal{D}_2$, $\mathcal{D}_3$;
% \IF{the Gaussian Mixture distribution is not symmetric}
% \STATE Compute $\widehat{\bfQ}_2$ using $\mathcal{D}_1$. Estimate the subspace $\widehat{\bfU}$   by orthogonalizing the eigenvectors with respect to the $K$ largest eigenvalues of  $\widehat{\bfQ}_2$
% \ELSE \STATE Pick an arbitrary vector  $\boldsymbol{\alpha} \in \R^{d}$, and use $\mathcal{D}_1$ to compute $\widehat{\bfQ}_3(\bfI_d,\bfI_d,\boldsymbol{\alpha})$.  Estimate   $\widehat{\bfU}$   by orthogonalizing the eigenvectors with respect to the $K$ largest eigenvalues of   $\widehat{\bfQ}_3(\bfI_d,\bfI_d,\boldsymbol{\alpha})$.
% \ENDIF 
\STATE Compute $\widehat{\bfQ}_2$ using $\mathcal{D}_1$. Estimate the subspace $\widehat{\bfU}$  with respect to the largest $K$ eigenvectors of $\widehat{\bfQ}_2$;
\STATE Compute $\widehat{\bfR}_3=\widehat{\bfQ}_3(\widehat{\bfU},\widehat{\bfU},\widehat{\bfU})$ from data set $\mathcal{D}_2$. Obtain $\{\hat{\bfv}_i\}_{i\in[K]}$ by decomposing $\widehat{R}_3$;
\STATE $\boldsymbol{\bar{w}}_i^* = \widehat{\bfU}\hat{\bfv}_i$ for $i\in[K]$;
\STATE Compute $\widehat{\bfQ}_1$ from data set $\mathcal{D}_3$. Estimate the magnitude $\widehat{\bfz}$ by solving the optimization problem
    $$\widehat{\boldsymbol{z}}= \arg\min_{\boldsymbol{\alpha}\in\mathbb{R}^K}\frac{1}{2}\|\widehat{\bfQ}_1-\sum_{j=1}^K \alpha_j\boldsymbol{\bar{w}}_j^*\|^2;$$
\STATE \textbf{Return:} $\hat{z}_j\widehat{\bfU}\hat{\bfv}_j$ as the $j$th column of $\bfW_0$,  $j\in [K]$.
\end{algorithmic}
\end{algorithm}
Subroutine \ref{TensorInitialization} estimates the direction and magnitude of $\bfw_j^*, j\in[K]$ separately. The direction vector is defined as $\boldsymbol{\bar{w}}_j^*:=\bfw_j^*/\|\bfw_j^*\|$, and the corresponding magnitude $\|\bfw_j^*\|$ is denoted as $z_j$.
Line 2 estimates  the subspace  $\widehat{\bfU}$ spanned by $\{\bfw_1^*,\cdots,\bfw_K^*\}$ using $\widehat{\bfQ}_2$.  Lines 3-4 estimate the  direction vector $\boldsymbol{\bar{w}}_j^*$ by employing the KCL algorithm \cite{KCL15}. Line 5 estimates  the magnitude $z_j$. Finally, the returned estimation of $\bfW^*$ is calculated as $\hat{z}_j\widehat{\bfU}\hat{\bfv}_j$.
\section{Theoretical results}\label{sec: theory}
% \subsection{Informal Theoretical Findings}
% The major insights of the major theorems as follows.

% \subsection{Formal Theorems}
Theorem 1 indicates that with sufficient number of samples as in \eqref{final_sp}, the iterates returned by proposed Algorithm \ref{gd} converge linearly to a critical point $\widehat{\bfW}_n$ near a permutation of the ground truth, denoted as $\bfW^*\bfP^*$. The distance between $\widehat{\bfW}_n$ and $\bfW^*\bfP^*$ are upper bounded as a function of the number of samples in \eqref{eqn:converge}. Additionally, the required number of samples depends on $\mathcal{\bfB}$, which is a function of input distributions in \eqref{notation_GMM}. 
\begin{thm}\label{thm1}
% Consider the binary classification problem with one-hidden-layer fully connected neural network as in (\ref{cla_model}). 
% Suppose  Assumption \ref{assumption1} holds, then there exist 
Given the samples from $(\mathcal{X},\mathcal{Y})$ 
with size $n$ satisfying
\begin{equation}
%\begin{aligned}
n \geq n_{\textrm{sc}}:=poly(\epsilon_0^{-1}, \kappa, K) \mathcal{B}(\boldsymbol{\lambda},\bfM) d\log^2{d}
  \label{final_sp}
%\end{aligned}
\end{equation}
for some $\epsilon_0\in(0,\frac{1}{4})$ and positive value functions $\mathcal{B}(\boldsymbol{\lambda},\bfM)$ and $q(\boldsymbol{\lambda},\bfM)$, with probability at least $1-d^{-10}$,
the iterates $\{\bfW_t\}_{t=1}^T$ returned by Algorithm \ref{gd} with step size $\eta_0=O\Big( \big(\sum_{l=1}^L\lambda_l(\|\tilde{\bfmu}_l\|_\infty+1)^2\big)^{-1}\Big)$  converge  linearly to a critical point $\widehat{\bfW}_n$ with the rate of convergence $v=1-K^{-2}q(\boldsymbol{\lambda},\bfM)$, i.e.,   
\begin{equation}
\begin{aligned}
 ||\bfW_t-\widehat{\bfW}_n||_F 
\leq  v^t||\bfW_0-\widehat{\bfW}_n||_F.
\label{linear convergence}
\end{aligned}
\end{equation}
Moreover, there exists a permutation matrix $\bfP^*$ such that the distance between $\bfW^*\bfP^*$ and $\widehat{\bfW}_n$ is bounded by
\begin{equation}\label{eqn:converge}
 ||\widehat{\bfW}_n-\bfW^*\bfP^*||_F \leq O\Big( K^{\frac{5}{2}}\cdot\sqrt{d\log{n}/n}\Big).
\end{equation}
\end{thm}
\textbf{Remark 1:} From \eqref{final_sp}, the required number of samples for successful estimation is a linear function of the input feature dimension $d$ and $\mathcal{B}(\boldsymbol{\lambda}, \bfM)$, where $\mathcal{B}$ is a function of the input distribution parameters. Note that the degree of freedom of $\bfW^*$ is $dK$, the sample complexity in \eqref{final_sp} is nearly order-wise optimal in terms of $d$. In addition, $\mathcal{B}(\boldsymbol{\lambda}, \bfM)$ is an increasing function of any entry of $\bfM$ (from Corollary 1 below), indicating that the sample complexity increases as the mean of any Gaussian component increases.

\textbf{Remark 2:} From \eqref{linear convergence}, the distance between the convergent point $\widehat{\bfW}_n$  and the ground truth $\bfW^*\bfP^*$ is in the order of $\sqrt{d\log n/n}$. When the number of samples increases, the convergent point moves closer to the ground truth, implying a smaller generalization error.
 
\textbf{Remark 3:} From \eqref{eqn:converge}, we can see that the iterations converge to $\widehat{\bfW}_n$ linearly, and the rate of convergence is $1-K^{-2}q(\boldsymbol{\lambda},\bfM)$, denoted as $\nu$. When the number of neurons decreases or the input distribution changes such that $q$ increases, the rate of convergence decreases, indicating a faster convergence.

% The sample complexity for estimation is  $\Theta(d\log^2{d})$, where $d$ is the feature dimension. Our bound is  almost order-wise optimal with respect to $d$ because the degree of freedom is $dK$. The additional multiplier of $\log^2{d}$ comes from the concentration bound in the proof technique. Note that we focus on the dependence of the sample complexity on the feature dimension $d$ and treat the network width $K$ as constant. Our result is in the same order as the sample complexity for  the standard Gaussian input in \cite{FCL20} and \cite{ZSJB17}, indicating that our technique can handle input from the Gaussian mixture model without increasing the order of the sample complexity. 

% \textbf{Remark 3:} 
% \textbf{3. Impact of  the mean}:
% If everything else is fixed, and at least 
% one entry of a mean $\mu_{l(i)}$ (the $i$th entry of $\bfmu_l$) of the Gaussian mixture model increases from 0 to infinity (in terms of the absolute value),  the sample complexity increases to infinity from a constant and the convergence slows down. The explanation is that when the absolute value of some mean increases, some training samples have a significantly large magnitude such that the sigmoid function saturates. The gradient of these samples is close to zero, so they are not informative for estimating $\bfW^*$. Therefore, the required samples complexity for estimation needs to increase,  and the gradient descent algorithm slows down. 

Corollary \ref{prop: diag_Cor} states the impact of the distribution parameters $\lambda$ and $\bfM$.
% $\mathcal{B}(\boldsymbol{\lambda},\bfM)$ and $q(\boldsymbol{\lambda},\bfM)$
% on the sample complexity $n_{\textrm{sc}}$ and the convergence rate $v$. 
Specifically, when the mean vectors $\{\bfmu_l\}_{l=1}^M$ in \eqref{notation_GMM} increases, the sample complexity $n_{\textrm{sc}}$ increases as $\mathcal{B}(\boldsymbol{\lambda},\bfM)$ increases, and the convergence rate $v$ increases as $q(\boldsymbol{\lambda},\bfM)$ decreases. Suppose some entry of $\bfmu_l$ go to infinity, $n_{\textrm{sc}}$ will go to infinity, and $\nu$ will converge to $1$.
\begin{corollary}\label{prop: diag_Cor} Recall $\bfM=(\bfmu_1,\cdots,\bfmu_L)$. Let $|\bfmu_{l(i)}|$  be the $i$-th entry of $\bfmu_l$,
\begin{enumerate}
    \item $\mathcal{B}(\boldsymbol{\lambda}, \bfM)$ is an increasing function of $|\bfmu_{l(i)}|$.
    \item $q(\boldsymbol{\lambda},\bfM)$ is a decreasing function of $|\bfmu_{l(i)}|$.
\end{enumerate} 
\end{corollary}

\section{Simulation}

Here we present our results for numerical experiments. All simulations are implemented in MATLAB 2021b on a workstation with 3.40GHz Intel Core i7. The weights of the teacher model $\bfW^*\in \R^{d\times K}$ are randomly generated such that each entry of $\bfW^*$ belongs to $\mathcal{N}(0,1)$. The corresponding training samples $\{\bfx_i, y_i\}_{i=1}^n$ are randomly selected following (\ref{notation_GMM}) and  (\ref{cla_model}) with the generated weights $\bfW^*$.  

\subsection{Sample complexity}
We first show how the number of required samples is related to the dimension of data. The parameters of input distribution are selected as $L=2$, $\lambda_1=\lambda_2=\frac{1}{2}$, $\bfmu_1=0.5*\boldsymbol{1}$ and $\bfmu_2=-\bfmu_1$. The number of neurons in the hidden layer is set as $3$. For a given $\bfW^*$, we initialize the starting point $M$ times randomly and denote $\widehat{\bfW}_n^{(m)}$ as the output of Algorithm \ref{gd} with the $m$-th initialization, and $M=20$. 
% The average estimation error $e_{\bfW^*}=\sqrt{\sum_{m=1}^M||\widehat{\bfW}_n^{(m)}-\bar{\bfW}_n||^2/M}$, where $\bar{\bfW}_n$ is the mean of $(\widehat{\bfW}_n^{(1)},\cdots, \widehat{\bfW}_n^{(L)})$. 
An experiment is viewed as a success if the average estimation error $e_{\bfW^*}$ is less than $10^{-3}$, where $e_{\bfW^*}$ is the standard deviation of $(\widehat{\bfW}_n^{(1)},\cdots, \widehat{\bfW}_n^{(L)})$.
  
\begin{figure}
\centering
\includegraphics[width=0.7\linewidth,height=0.5\linewidth]{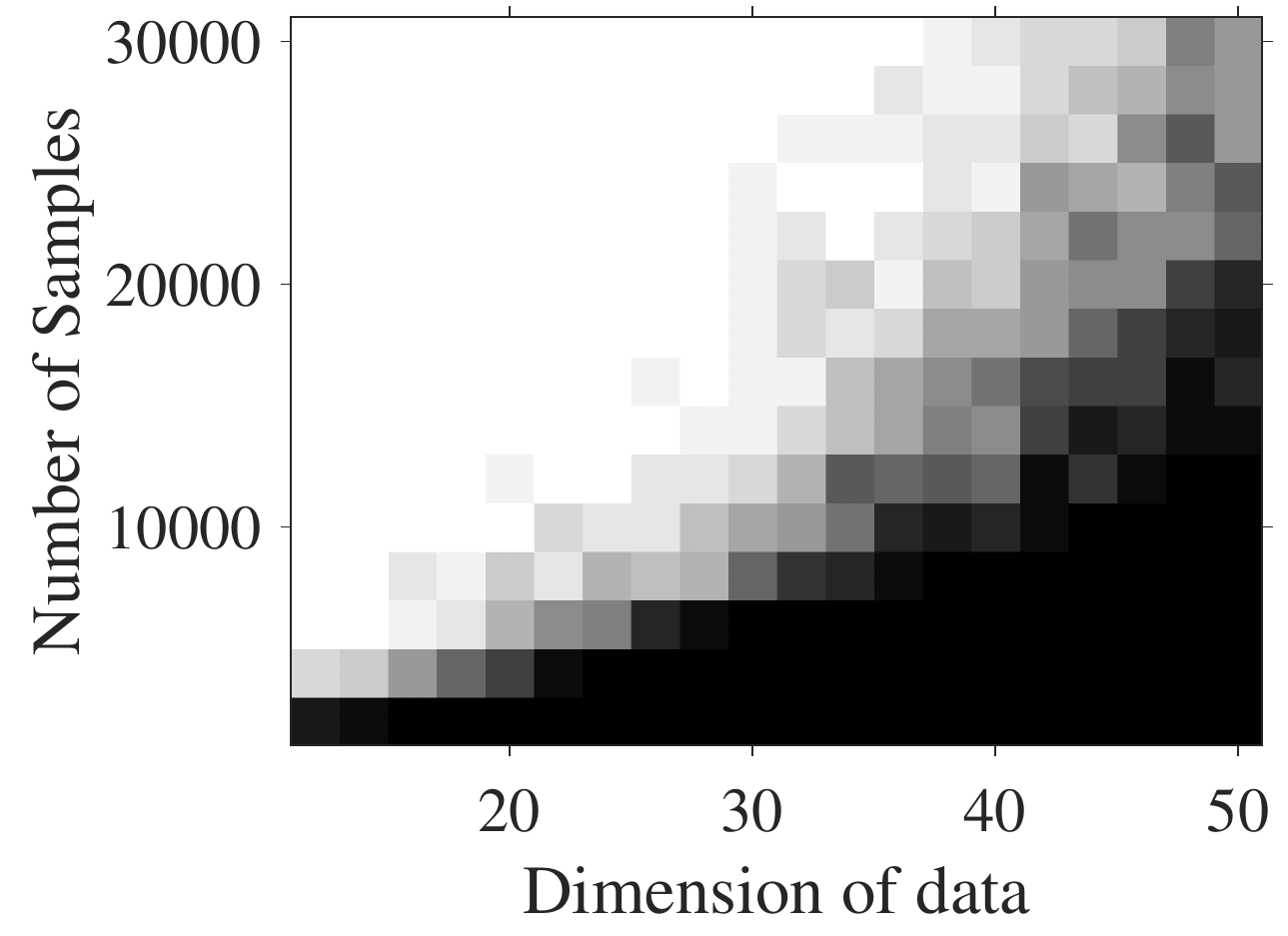}
\caption{The sample complexity against the feature dimension $d$}
\label{fig:ntod}

\end{figure}

In Fig.~\ref{fig:ntod},  we increase the feature dimension $d$ from 12 to 50 by 2 and vary the number of samples $n$ from $2\times 10^3$ to $3\times 10^4$. We test $20$ independent experiments for each pair of $d$ and $n$ and then show the average success rate using grey blocks, where black ones mean rate 0 and white ones mean rate 1. In Fig.~\ref{fig:ntod}, the boundary line of black and white parts is almost straight, indicating an approximate linearity between $n$ and $d$, which verifies our result in \eqref{final_sp}.

\begin{figure}
\centering
\includegraphics[width=0.75\linewidth]{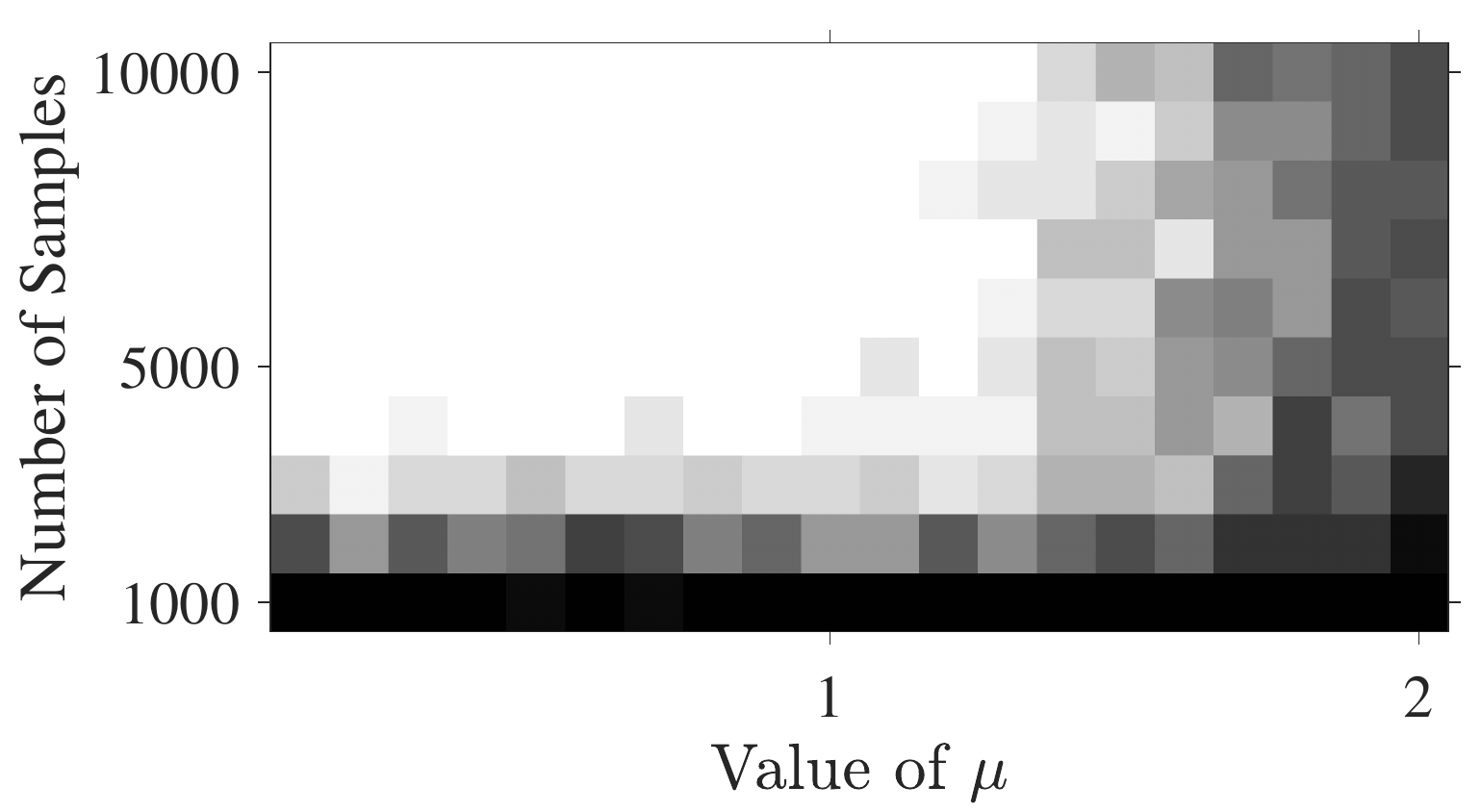}
\caption{The sample complexity when one mean changes}
\label{figure: spmu}

\end{figure}

We then study the impact on the sample complexity when the mean value in the Gaussian mixture model changes. Set $d=10$, $\lambda_1=0.4$, $\lambda_2=0.6$, $\bfmu_1=\mu \cdot\boldsymbol{1}$, $\bfmu_2=\boldsymbol{0}$. Fig.~\ref{figure: spmu} shows that when the mean increases from $0$ to $2$, the sample complexity increases, matching our theoretical analyses in Section \ref{sec: theory}. 

\subsection{Convergence analysis}
\begin{figure}
\centering
\includegraphics[width=0.7\linewidth,height=0.4\linewidth]{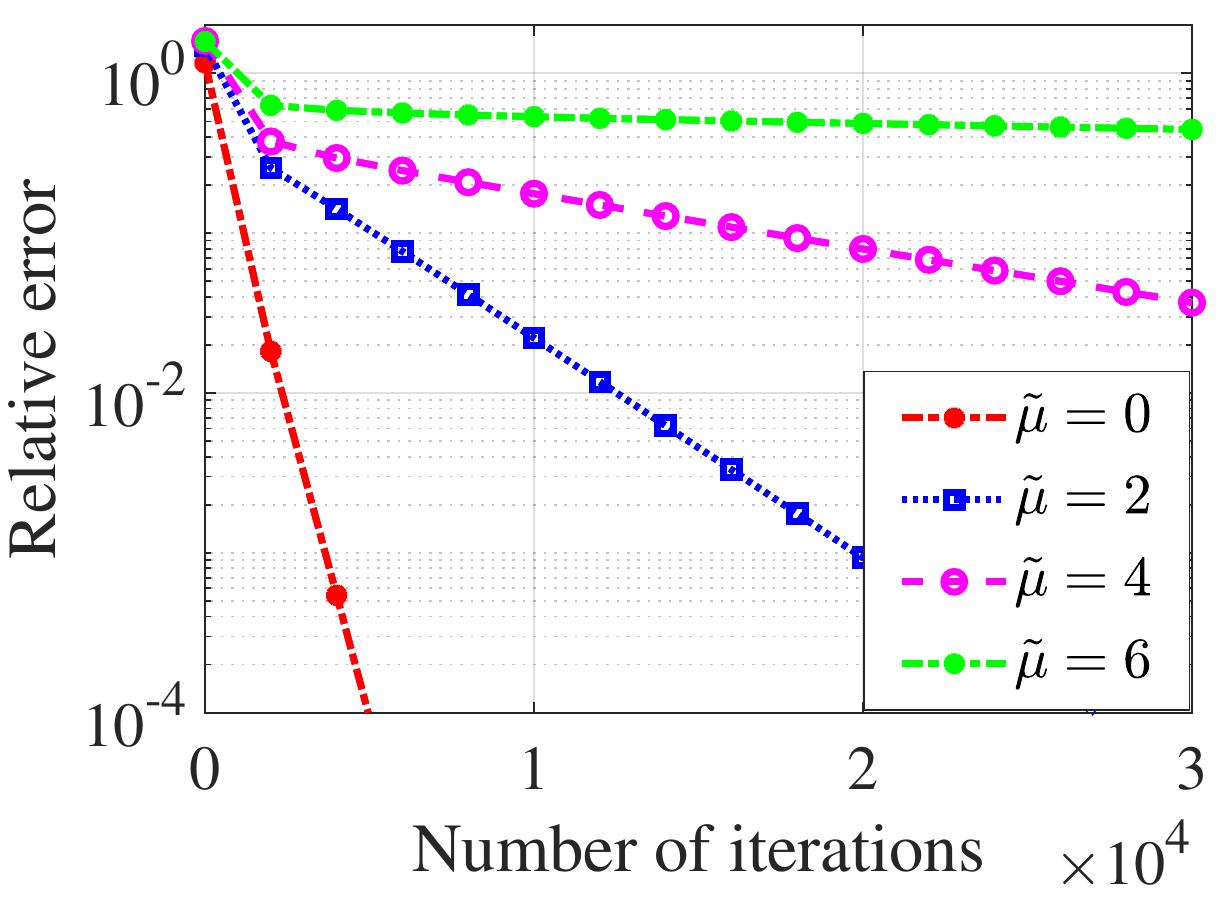}
\caption{The convergence rate with different $\tilde{\mu}$}
\label{figure: convergence_musigma}

\end{figure}

We next fix $d=5$ and evaluate the convergence rate of Algorithm \ref{gd}. 
We set $n=1\times 10^4$, $\lambda_1=\lambda_2=0.5$, $\bfmu_1=-\bfmu_2=C\cdot\boldsymbol{1}$ for a positive $C$. We vary $C$ and let $\tilde{\mu}=\max_{l}\|\tilde{\bfmu_l}\|_\infty$. Recall the relative error is defined as $\|\bfW_t-\widehat{\bfW}_n\|_F$ in \eqref{linear convergence}. Fig.~\ref{figure: convergence_musigma} shows the linear convergence of Algorithm \ref{gd} for different $\tilde{\mu}$ and the impact of $\|\tilde{\bfmu}_l\|_\infty$. As predicted in Theorem \ref{thm1},  when $\tilde{\mu}$ increases, gradient descent converges slower.

\section{Conclusion}
In this paper, we study the problem of learning a fully connected neural network when the input features belong to the Gaussian mixture model from the theoretical perspective. We propose a gradient descent algorithm with tensor initialization, and the iterates are proved to converge linearly to a critical point with guaranteed generalization. Additionally, we establish the sample complexity for successful recovery, and the sample complexity is proved to be dependent on the parameters of the input distribution. Possible future direction is to analyze the influence of variance on the learning performance.

% \section*{Acknowledgment}

% The preferred spelling of the word ``acknowledgment'' in America is without 
% an ``e'' after the ``g''. Avoid the stilted expression ``one of us (R. B. 
% G.) thanks $\ldots$''. Instead, try ``R. B. G. thanks$\ldots$''. Put sponsor 
% acknowledgments in the unnumbered footnote on the first page.

%\bibliographystyle{IEEEtran}
%\bibliography{refer/ref, refer/ref_Hongkang, refer/MengWangPub, refer/self}

\end{document}